\documentclass{article}
\usepackage{amsfonts}
\usepackage{url}
\usepackage{amssymb}
\usepackage{amsmath}
\newcommand{\claimedsubsumes}{{\bf claimed\_subsumes}}
\newcommand{\claimedrefutes}{{\bf claimed\_refutes}}
\newcommand{\newer}{{\bf more\_recent}}

\newtheorem{example}{Example}
\newtheorem{definition}{Definition}
\begin{document}

\title{Recommending the Most Encompassing Opposing and Endorsing Arguments in Debates}
\author{Marius C. Silaghi and Roussi Roussev\\
Florida Institute of Technology\\
\url{{msilaghi,rroussev1998}@fit.edu}}
\maketitle

\begin{abstract}
  Arguments are essential objects in DirectDemocracyP2P, where they
  can occur both in association with signatures for petitions, or in
  association with other debated decisions, such as bug sorting by
  importance.

  The arguments of a signer on a given issue are grouped into one
  single justification, are classified by the {\em type} of signature (e.g.,
  supporting or opposing), and can be subject to various types of threading.

  Given the available inputs, the two addressed problems are: (i) how
  to recommend the best justification, of a given type, to a new
  voter, (ii) how to recommend a compact list of justifications
  subsuming the majority of known arguments for (or against) an
  issue.
  We investigate solutions based on weighted bipartite graphs.
\end{abstract}
\section{Introduction}

The process of gathering signatures for petitions has been extended by
enabling supporters and opposers to add justifications for their
stances (\url{DirectDemocracyP2P.net}). As a peculiarity of this
system, authors of the signatures are
authenticated, and each author can reference a single justification
for a given petition.  Given the large number of justifications that
can be submitted in such a system, the need for a recommender system to
help spot the most relevant justifications is emerging.

\paragraph{Opposition and abstention.}
While commonly petitions can be signed only for support, we assume
that that an electronic system can afford and benefit from {\em
  enabling manifestations of opposition or abstention} to a petition.
These other manifestations can offer observers a hint to the strength
of the opposition to the petition (even if less statistically relevant
than polls with a controlled distribution of the subjects). More
importantly, these other manifestations provide a mechanism enabling
people not supporting the petition to provide early a
controlled/authenticated feedback to the petition supporters and
observers, via justifications as described next.

\paragraph{Justification.}
We also consider it important to {\em enable the submission of
  justifications} (i.e. some natural language explanation) with the
signatures of support or opposition. A justification contains a set of
arguments. These justifications can be seen as ways to boost reciprocal
understanding between people with opposing views, and as a way to
encourage activism~\cite{Hibbing13}. They can also help bring democracy in media,
mitigating its frequently raised critique of manipulation and
censorship by powerful groups of interests.

\paragraph{Justification Type.} A justification can be classified based on
the type of signature (i.e., support, opposition, abstention) that
it accompanies. Here we start by proposing solutions with only two
types of justification: {\em supporting justification} and {\em
opposing justification}. Two justifications are of the same type
if both accompany the same type of signature (support or opposition).

\paragraph{Threading: Relations between justifications.} Unlike
classical argumentation where relations are extracted from formal
arguments, we assume that certain relations are explicitly offered by
associating them with opaque arguments (in an opaque justification).
This is commonly done in existing fora, where relations are provided
via a threading model (e.g., each comment may answer another comment).
While a formal logical argumentation could be used as support for much
more complex mechanisms, the mechanism of opaque arguments we use can
be seen as a basic case, where arguments in each provided
justification form the premise of the associated vote (support or
opposition):
$$arguments \rightarrow petition$$
$$arguments \rightarrow \neg petition$$
The simplification to this basic case can help us to concentrate
better on the semantic of the conclusion (i.e., associated signature).
This is an element that was not sufficiently analyzed in the past.
Once this semantic is well established (with appropriate weights given
to the relations), the extension to more complex arguments (i.e., to
restricted languages) can be seen as a combination with what has been
done in the past in the argumentation area.

Another type of relation we support is {\bf contradicts}, where a justification
is presented as an answer to a different justification that it
corrects or enhances.
A third type of relation that we discuss is {\bf includes}, claiming that a
justification subsumes a second justification (claim explicitly introduced
by a voter and that may or may not be automatically verifiable).
The last type of relation is {\newer} and can sort justifications based on their submission date.

In the following, this approach/interface to electronic petitions
(i.e., with opposing views and justifications) is considered a given.
Other approaches may be possible and further work may generate and
compare such competing approaches to electronic petitions.

\section{Background and Problem}
Work in {\em argumentation framework} focuses on multiple final goals:
\begin{enumerate}
\item
finding a set of laws (rules) that are compatible and have support~\cite{Dokow}.
\item
finding the strongest chains of arguments~\cite{Amgoud}.
\item
aggregating arguments into a summary of arguments~\cite{Pigozzi}.
\end{enumerate}
All these efforts 
refer to the case when a committee must follow some rules (being
supervised by a superior authority).  This superior authority wants to
understand the bases of the decisions and to verify them.  They start
with a set of values and prove that they can logically justify their
decisions based on these values (not being accusable of corruption).

In our case this is not applicable since in our setting (a kind of
forum) everybody has a different set of values. Therefore on a free
forum the arguments are in general not comparable, and the generation
of a unique justification (proving lack of corruption) is not the main
issue at this point.  As such, here we assume that in petitions, the
declared utility (i.e. type of signature) is more important than the
logic correctness of arguments.  On politically sensitive topics a
human frequently uses arguments to justify a sentimentally taken
decision (often inherited from her social circle), decision taken
prior to the crystallization of her own arguments~\cite{Hibbing13}.  

\begin{example}
For example, if one
succeeds to show a person from a different political party that his
arguments are false, this commonly does not make the person to change
his opinion but rather to search and find new arguments (as can be
observed in the US from generations of debating republicans and democrats
that do not seem to converge)~\cite{Alford13}.
\end{example}
Arguments may help people to understand and tolerate each other (in so
much as this understanding removes fear)~\cite{Alford13,Hibbing13},
but from a democratic perspective, what is important is the declared
utility (the taken decision/vote), more than the arguments.

In DirectDemocracyP2P, we have a somewhat different framework (both
from the perspective of the environment, and in what is desirable and
expected as a result of the study of arguments):
\begin{enumerate}
\item[A.] We assume no supreme authority individual, but rather the sole supreme
  authority is formed by the entire constituency (at least in {\em grassroot
    organizations}), and arguments are presented in natural language.
  Since the arguments are in natural language we do not expect that
  they can be robustly parsed and prepared as formal logical
  statements. Therefore the only thing that we plan to exploit automatically is
  the association vote-justification and the structure obtained from threading.
\item[B.] Therefore, following (A), the final goal of an argumentation
  study would be the help given to a
  constituent in finding the most complete proposed arguments, as
  revealed by her predecessors' inputs. This is, we should formalize a partial
  argumentation (under development).  A voter (new or ancient) studies
  current arguments with the purpose of selecting the most complete
  justification for her vote (justification which can be subjective).
\end{enumerate}

We address two problems: finding candidate justifications, and suggesting components 
of a new justification.
\paragraph{Problem 1: Find candidate justifications.}
If an existing justification subsumes all the arguments that a new
voter considers relevant (from this voter's point of view), then the
voter will select that particular justification. Therefore we want an
algorithm that detects candidates for such a {\em subsuming
  justification} to be used as someone's justification of a binary
conclusion (pro or against the petition). The overall system can be
seen as a Captcha exploiting social computation to find subsuming
justifications to a current pool of arguments (found in a set of
justifications).

When no such justifications exists, that according to a new voter
subsumes all known arguments from the point of view of the voter, then the
voter will create a new justification.

The detection of the subsuming justification will be based on a
hierarchy of justifications. We cannot build this hierarchy on the
base of the arguments found in the content of the justification, as we
agreed to consider this content as being opaque. However we can build
it on the basis of some relations between justifications.

We can have two types of (voted) relations between two justifications $a$ and $b$.
\begin{enumerate}
\item
a {\claimedsubsumes} b (read: justification a {\bf is claimed to subsume} justification b).
\item
a {\claimedrefutes} justification b (where b has an opposite conclusion/vote to a, read: justification a {\bf is claimed to refute} justification b).
\end{enumerate}

The concepts of {\claimedsubsumes} and {\claimedrefutes} can be defined from
the perspective of logic or social and political sciences. Finally the
definition to be used is to be left to the users specifying these
relations (since the natural language arguments are anyhow opaque to
the computational system, which cannot verify and enforce a definition).
For generality, for now we treat these relations simply as arcs in a
bipartite graph. This bipartite graph can be exploited in ways that
fit the expectations of the users.

What we can do, is an algorithm using available inputs to propose the
{\em best} (i.e., most complete) subsuming candidates for
justifications of the two opposing conclusions.

\paragraph{Problem 2: Suggesting components of a new justification.}
Another problem that is raised in our framework is: {\em What is the
  ``best'' subset that subsumes the arguments of a conclusion/vote
  (based on the two types of voted relations). This is necessary not
  only for the voter that wants to propose a new argument, subsuming
  all the old ones, but also for the undecided voter that wants to
  study existing arguments in order to construct her opinion.}

In order to measure ``good'' in this problem, as well as in the previous problem, we can propose several metrics:
\begin{enumerate}
\item
the most voted
\item
the newest (in time)
\item
the most complete (from the perspective of the existing relations)
\end{enumerate}
Further, metrics can consist of any combinations of these three ones proposed here.

\section{Concepts}
Here, after defining the concept of {\em answer}, we introduce
incrementally three basic frameworks (used by the two algorithms in the
subsequent section). Then one can define generalizations and
combinations of these frameworks.

\begin{definition}[Answer]
  A justification is said to {\bf answer} to a {\em voter} if either
  it is associated with the signature of that voter, or if it was
  created with a specification that it \claimedrefutes{} the
  justification selected by that voter.
\end{definition}

\begin{definition}[Subsuming Justification Problem (SJP)]
  The Subsuming Justification Problem (SJP) for a given petition $M$
  consists of a tuple $\langle N, P, V, R, K \rangle$.
  Here $N=\{n_1,....,n_{m_n}\}$ is a set consisting
  of $m_n$ opposing justifications of $M$, and $P=\{p_1,...,p_{m_p}\}$
  is a set of $m_p$ supporting justifications for $M$.

  Each justification $j$ is associated with a number of $v_j$
  signatures, as per the set $V=\{(j,v_j)\mid j\in N\cup P, v_j=signatures(j)\}$.  The relation $R:P\cup S\rightarrow {\cal P}(P\cup S)$ where
  $R\mid_P:P\rightarrow N$ and $R\mid_N:N\rightarrow P$,
  associates each opposing justification $n_i$ with at most one
  supporting justification $p_{n^i}$, and each supporting
  justification $p_i$ with at most one opposing justification
  $n_{p^i}$, by the \claimedrefutes{} relation.
\begin{eqnarray*}
n_i &\xrightarrow{v_{n_i}}& \neg M\\
p_i &\xrightarrow{v_{p_i}}& M\\
p_i &\claimedrefutes& n_{p^i}\\
n_i &\claimedrefutes& p_{n^i}.
\end{eqnarray*}

The SJP problem is to find a set of at most $K$ supporting
justifications that {\bf answer} to a maximum number of signatories
(both supporting and opposing $M$), and a set of at most $K$
opposing justifications that {\bf answer} to a maximum number of signatories
(either supporting and opposing $M$).
\end{definition}

In a further complication it is possible to have each voter specify
explicitly the justification that his selected justification
\claimedrefutes{} (rather than inheriting the one specified at the
creation of his justification). This allows to better adjust
the relations from good old justifications to \newer{} justifications.

\begin{definition}[Weighted Subsuming Justification Problem (WSJP)]
  The Weighted Subsuming Justification Problem (WSJP) for a given petition $M$
  consists of a tuple $\langle N, P, V, R, K \rangle$.
  Here $N=\{n_1,....,n_{m_n}\}$ is a set consisting
  of $m_n$ opposing justifications of $M$, and $P=\{p_1,...,p_{m_p}\}$
  is a set of $m_p$ supporting justifications for $M$.

  Each justification $j$ is associated with a number of $v_j$
  signatures, as per the set $V=\{(j,v_j)\mid j\in N\cup P,
  v_j=signatures(j)\}$.  The relation $R \subset (N\times \mathbb{N}\times P) \bigcup (P \times
  \mathbb{N}\times N) $ associates a weight to
  each pair between an opposing justification $n_i$ and supporting
  justification $p_{j}$, and to each pair between a supporting
  justification $p_i$ and an opposing justification $n_{j}$, by the
  \claimedrefutes{} relation.  Each element $(i,w_{i,j},j)$ of the $R$
  relation is weighted with the number $w_{i,j}$ of signatories of the
  left-hand justification $i$ that have explicitly stated that this
  justification \claimedrefutes{} the justification $j$ on the right-hand
  of the relation.
\begin{eqnarray*}
n_i &\xrightarrow{v_{n_i}}& \neg M\\
p_i &\xrightarrow{v_{p_i}}& M\\
p_i &\overset{w^p_{i,j}}{\claimedrefutes}& n_j\\
n_i &\overset{w^n_{i,j}}{\claimedrefutes}& p_j.
\end{eqnarray*}

The WSJP problem is to find a set of at most $K$ supporting
justifications that {\bf answer} to a maximum number of signatories
(both supporting and opposing $M$), and a set of at most $K$
opposing justifications that {\bf answer} to a maximum number of signatories
(either supporting and opposing $M$).
\end{definition}

To detect a reduced set of justifications that cover existing
arguments, to be used in creating a new justification, one has to
consider the \claimedsubsumes{} relation.

\begin{definition}[Components Subsuming Justification Problem (CSJP)]
  The Components Subsuming Justification Problem (CSJP) for a given petition $M$
  consists of a tuple $\langle N, P, V, R, S, K \rangle$.
  Here $N=\{n_1,....,n_{m_n}\}$ is a set consisting
  of $m_n$ opposing justifications of $M$, and $P=\{p_1,...,p_{m_p}\}$
  is a set of $m_p$ supporting justifications for $M$.

  Each justification $j$ is associated with a number of $v_j$
  signatures, as per the set $V=\{(j,v_j)\mid j\in N\cup P, v_j=signatures(j)\}$.  
  The function $R:P\cup S\rightarrow {\cal P}(P\cup S)$ where
  $R\mid_P:P\rightarrow N$ and $R\mid_N:N\rightarrow P$,
  associates each opposing justification $n_i$ with at most one
  supporting justification $p_{n^i}$, and each supporting
  justification $p_i$ with at most one opposing justification
  $n_{p^i}$, by the \claimedrefutes{} relation.
  The function $S:P\cup S\rightarrow {\cal P}(P\cup S)$ where
  $S\mid_P:P\rightarrow {\cal P}(P)$ and $S\mid_N:N\rightarrow {\cal
    P}(N)$, associates each justification $j$ to a set of
  justifications of the same type that it \claimedsubsumes{}.
\begin{eqnarray*}
n_i &\xrightarrow{v_{n_i}}& \neg M\\
p_i &\xrightarrow{v_{p_i}}& M\\
p_i &{\claimedrefutes}& n_{p_i}\\
n_i &{\claimedrefutes}& p_{n_i}\\
j &\claimedsubsumes &k, \forall k \in S(j)
\end{eqnarray*}

The CSJP problem is to find a set of at most $K$ supporting
justifications that {\bf answer} to a maximum number of signatories
(both supporting and opposing $M$), and a set of at most $K$
opposing justifications that {\bf answer} to a maximum number of signatories
(either supporting and opposing $M$).
\end{definition}

\section{Algorithms}
Each of these frameworks (as well as their combinations) can be
solved approximately via an algorithm {\bf similar} to mini-max, that
traverses the search tree down to a certain depth. More exactly, in the basic
case one starts with the given justification and in subsequent steps one
can apply {\bf a kind of} transitivity of the relation \claimedrefutes{}.
Under the assumption that each voter selects the most complete justification
fitting his vote, this transitivity is of the type:
\begin{eqnarray*}
p ~\claimedrefutes{}~ n' \\ 
n' ~\claimedrefutes{}~ p' & \rightarrow & p ~\claimedrefutes{}~ n \\ 
p' ~\claimedrefutes{}~ n \\ 
\end{eqnarray*}
Using this special transitivity one can search for the justifications
that (within a limited depth) refute the largest number of
justifications of the other type.

The algorithm pseudo-code is shown in Figure~\ref{fig:co}:
\begin{figure}
\fbox{
\parbox[c]{12cm}{
\begin{enumerate}
\item[]Function: refutes(j, level)
\item[]{\bf for} any $i$ s.t. $j$ \claimedrefutes{} $i$, add $i$ to $R_1$
\item[]{\bf for} $(k=2;k\leq{}level;k++)$\\
 $~~~~~R_k=\{i \mid u\in R_{k-1}, \exists t, u~\claimedrefutes{}~t \wedge t~\claimedrefutes{}~i\}$
\item[]{\bf return} $\bigcup_{k=1}^{level} R_k$
\end{enumerate}
}
}
\caption{Algorithm to find best counter-arguments}\label{fig:co}
\end{figure}

One can integrate the votes on justifications (and relations) as weights
to arguments (modeling their importance), and they can further be discounted
with a factor $\gamma<1$ to consider their depth in the tree:
$$\sum_{k=1}^{level} \gamma^{k}*votes(k)$$
where $votes(k)$ can integrate the number of signatures
for all justifications at level $k$ as well as for the \claimedrefutes{}
relations in the two directions used for the transitivity:
$votes(k)=votes_j(k)+\alpha*votes_{\rightarrow}(k)+\beta*votes_{\leftarrow}(k)$

This function can be applied to all justifications (for some level),
and then one can compute the cardinality of the results to estimate the
justifications containing the most arguments.
$$J_{level}:=\max_{j}\mid refutes(j,level) \mid$$
The higher the level, the worse propagate errors from the mentioned
assumptions. The lower the level, the less insight is available into
the debate. We foresee that the bests levels will be somewhere in the
set $\{2,3,4\}$. The best parameters will be identified as described later (based
on a simulation).

Any algorithm to compute a {\bf transitive closure} can exploit the
\claimedsubsumes{} relations to define a closure
(a small set of justifications subsuming most other relevant
justifications).

From the perspective of graph theory, the analysis can be done with 
bipartite graphs (supporting justifications vs. opposing
justifications).

Combinations and generalizations can handle the fact that the static relations from
framework SJP and CSJP can be voted individually (as at framework WSJP), etc.

The frameworks can further be augmented by association with a timestamp to
each argument. Thereby one obtains an additional relation, {\newer},
that enables an extension of the concept {\bf answers}.

\paragraph{Generative models}
It is also possible to build a Bayesian network modeling this
problem. Let us denote with $A_M$ the set of possible arguments for the 
petition $M$. Each justification can contain any subset of $A_M$,
and ca be modeled either with a discrete or a continuous variable.

In such a network corresponding to a SJP, there are two
nodes, $A_j$ and $R_j$, for each justification $j$, introduced in the
network in the order induced by the {\newer} relation.
The domain of $A_j$ is the power set of $A_M$, ${\cal P}(A_M)$.
The domain of $R_j$ is the set of possible justification (that are 
less recent than $j$), and specifies a justification
that $j$ \claimedrefutes{}.

The distribution of $A_j$ is uniform over the power set of $A_M$.  We
could consider that the nodes $A_j$ are independent\footnote{Or alternatively that
  it is restricted to justifications subsuming {\newer}
  justifications.}, while the nodes $R_j$ are dependent on all $A_i$
that correspond to justifications $i$ that are \newer{} than $j$.
The distribution of $R_j$ assigns to prior justifications a
probability proportional with the number of arguments that they
contain.
The votes could be distributed proportionally with the number of arguments.

The $A_j$ variables are hidden while the $R_j$ variables are evidence.

The number of values for the variables of such a Bayesian Network is
pretty high, therefore more research is needed on how to efficiently
exploit it directly for inferring the the justifications where $A_j$
are assigned to the largest subsets of $A_M$.

This network can be used for generating (sampling) test cases, as an
artificial ``ground truth'' (valid in as far as it correctly models
the world). Theoretically, Bayesian networks are general enough to
model quite complex human behavior. This potential ``ground truth''
can be used for evaluating the success of the recognition with the
previous algorithms (mini-max for justifications).

\paragraph{Deliberation}
In the area of argumentation, researchers strive to have computers
deliberate on behalf of humans, reasoning with logical arguments.
In our case (since each participant has different foundational values,
the deliberation is not something automated by artificial intelligence.
Rather, the deliberation is made by the people (who reevaluate their 
fears based on seen arguments), while the artificial intelligence is
used to help people find the most relevant arguments.

\section{Conclusion}
We propose a logical framework to reason about arguments in debates.
It is a type of abstract argumentation framework and can be used to
recommend relevant opinions to readers navigating the debate graph.

The framework is particularly relevant for debates where arguments
(seen as atomic entities in an attack/support relation) are submitted
in a decentralized fashion by a set of involved peers. 
This framework is designed as a mechanism to provide recommendations
in the \url{DirectDemocracyP2P.net} system.

\paragraph{Acknowledgments}
We acknowledge the help of Dr. Ioan Alfred Le\c{t}ia 
who recommended us the relevant related work in the argumentation area.

\bibliographystyle{plain}
\bibliography{arguments.bib}

\end{document}